\documentclass[lettersize, 10 pt, journal, twoside]{IEEEtran}

\IEEEoverridecommandlockouts     

\usepackage{color}
\usepackage[utf8]{inputenc}
\usepackage{amsmath,amssymb}
\usepackage{mathtools}
\usepackage{amsthm}
\usepackage{graphicx}
\usepackage{float}
\usepackage{wrapfig}
\usepackage{algorithmicx,algpseudocode}
\usepackage{algorithm}
\usepackage{amsfonts,dsfont}
\usepackage{mathrsfs}  
\usepackage{booktabs}
\usepackage{hyperref}
\usepackage{cite}

\newcommand{\norm}[1]{\lVert #1 \rVert}

\renewcommand{\phi}{\varphi}
\renewcommand{\epsilon}{\varepsilon}

\usepackage{hybridsystem}

\DeclareMathOperator*{\argmin}{argmin}

\newcommand{\fmatrix}[1]
{\begin{bmatrix} f{#1}^{r}
(x #1)\\ 
f{#1}^{s}
(x #1) \end{bmatrix}}
\newcommand{\gmatrix}[1]{\begin{bmatrix} g{#1}^{r}_1
(x {#1}) & g{#1}^{r}_2
(x {#1}) \\
    0   
    & g{#1}^{s}
    (x {#1})   \end{bmatrix}}

\newcounter{thm}
\newcounter{lem}
\newcounter{cor}
\newcounter{prop}
\newcounter{defn}
\newcounter{rem}
\begin{document}

\title{Powered Prosthesis Locomotion on Varying Terrains: Model-Dependent Control with Real-Time Force Sensing}

\author{Rachel Gehlhar$^{1}$, Je-han Yang$^{2}$, and Aaron D. Ames$^{1}$ \textit{Fellow, IEEE}%
\thanks{Manuscript received: September 9, 2021; Revised January 14, 2022; Accepted February 7, 2022.}
\thanks{This paper was recommended for publication by Editor Pietro Valdastri upon evaluation of the Associate Editor and Reviewers' comments.
This work was supported by the NSF GRF under Grant No. DGE‐1745301, the NSF Awards 1923239 and 1924526, and Wandercraft under Award No. WANDERCRAFT.21. This research was approved by California Institute of Technology Institutional Review Board with protocol no. 21-0693 for human subject testing.}
\thanks{$^{1}$Rachel Gehlhar and Aaron D. Ames are with the Department of Mechanical and Civil Engineering, J. Yang is with the Department of Electrical Engineering, California Institute of Technology, Pasadena, CA 91125 USA
        {\tt\footnotesize rgehlhar@caltech.edu, ames@caltech.edu}}%
\thanks{$^{2}$ Je-han Yang is with Department of Electrical Engineering, California Institute of Technology, Pasadena, CA 91125 USA
        {\tt\footnotesize jyang5@caltech.edu}}%
\thanks{Digital Object Identifier (DOI): see top of this page.}
}

\markboth{IEEE Robotics and Automation Letters. Preprint Version. Accepted February, 2022}
{Gehlhar \MakeLowercase{\textit{et al.}}: Powered Prosthesis Locomotion on Varying Terrains}

\maketitle

\begin{abstract}
Lower-limb prosthesis wearers are more prone to falling than non-amputees. Powered prostheses can reduce this instability of passive prostheses. While shown to be more stable in practice, powered prostheses generally use model-independent control methods that lack formal guarantees of stability and rely on heuristic tuning. Recent work overcame one of the limitations of model-based prosthesis control by developing a class of provably stable prosthesis controllers that only require the human interaction forces with the prosthesis, yet these controllers have not been realized with sensing of these forces in the control loop. Our work realizes the first model-dependent prosthesis knee controller that uses in-the-loop on-board real-time force sensing at the interface between the human and prosthesis and at the ground. The result is an optimization-based control methodology that formally guarantees stability while enabling human-prosthesis walking on a variety of terrain types.  Experimental results demonstrate this force-based controller outperforms similar controllers not using force sensors, improving tracking across 4 terrain types.
\end{abstract}


\IEEEpeerreviewmaketitle

\section{Introduction}
\IEEEPARstart{L}{ower}-limb prosthesis users fall more frequently than non-amputees \cite{AtlasAmptLimbDeficiencies}. A survey in \cite{AmputeeFalls} found 45\% of polled amputees had fallen in the past year while wearing their prosthesis. This instability could be due to their passive prostheses, which can be less stable than powered prostheses \cite{ClinicalComparison}. Current powered prosthesis control methods tend to be model-independent \cite{DesignControlTransProsth, VirtConsCtrlProst}, require heuristic tuning, lack formal guarantees of stability, and do not utilize the system's natural dynamics and force interactions with the user and environment. Current state of the art controllers tune impedance parameters for multiple discrete phases of a gait cycle for different subjects and behaviors \cite{goldfarb2021stair, AnnSimonConfig5, herr2021design}. Model-based control methods hold potential to yield a more transferable method between devices, users, and behaviors since they rely on measurable model parameters and inputs instead of a large set of heuristic tuning parameters. Additionally, model-based methods can be designed to yield formal guarantees of stability since they are based on the actual system dynamics. This motivates developing model-based prosthesis control methods that lend a more transferable method between applications, and guarantee stability for the user.

A challenge arises in developing model-dependent prosthesis controllers: \emph{the human dynamics are unknown}. The work of \cite{gehlhar2021separable} addressed this limitation by developing rapidly exponentially stabilizing control Lyapunov functions (RES-CLFs), which were shown to stabilize bipedal robotic walking \cite{ames2014rapidly}, in the context of separable systems \cite{StableRobustHZD,gehlhar2019control}. This resulted in a class of stabilizing prosthesis controllers relying only on local prosthesis information. Previously, RES-CLFs were difficult to realize on hardware due to the typical required inversion of the inertia matrix which is computationally expensive and susceptible to modeling error. The work of \cite{reher2020inverse} developed and demonstrated a RES-CLF controller on a bipedal robot without inverting the inertia matrix by constructing the RES-CLF in an inverse dynamics framework, realized as a quadratic program (QP) \cite{Bjrck1996NumericalMF}. The work of \cite{reher2020inverse} brings the class of controllers developed in \cite{gehlhar2021separable} closer to being implementable.

\begin{figure} [t] 
\centering
\includegraphics[width=1\columnwidth]{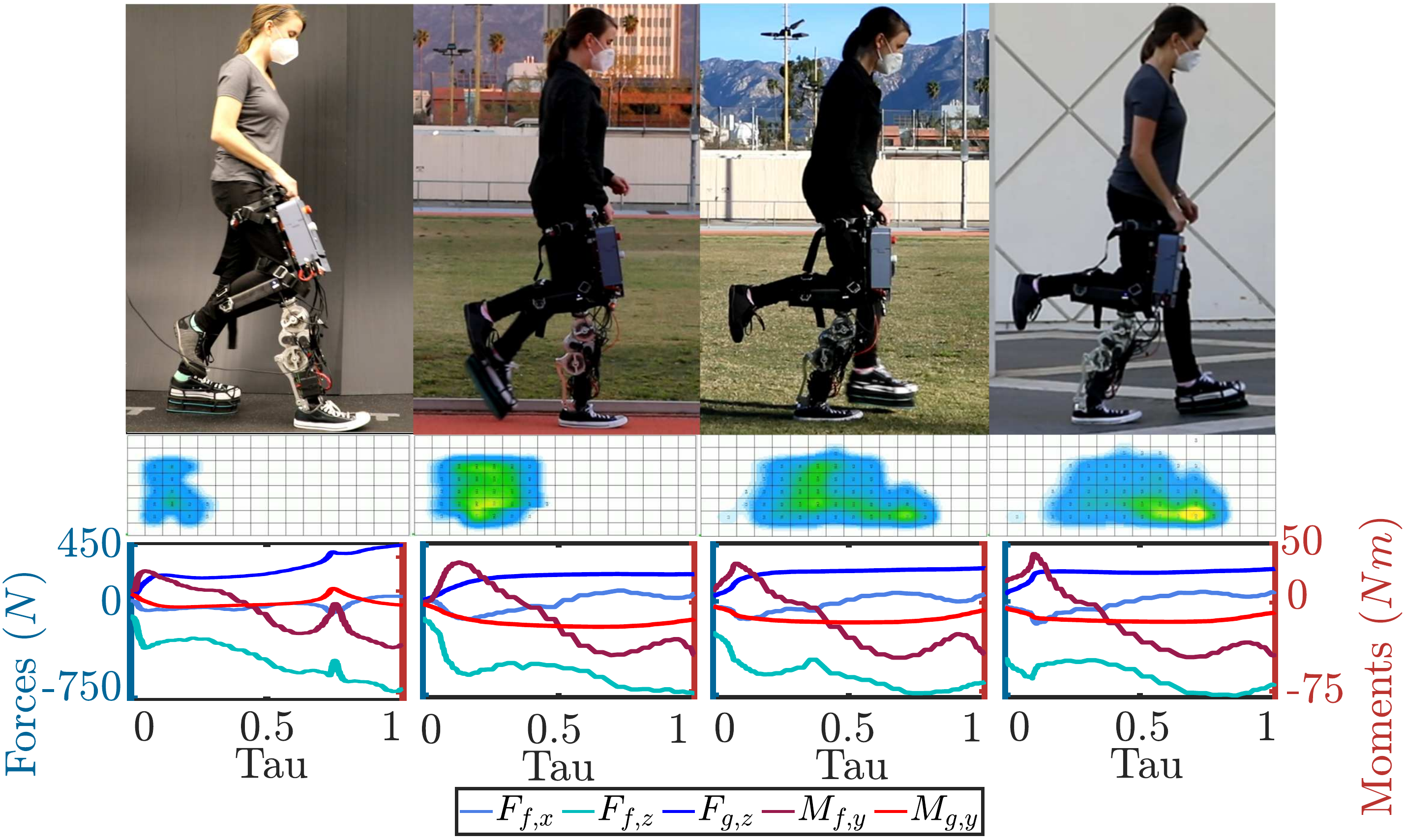}
\vspace{-0.75cm}
\caption{(top) Gait tiles of human subject walking with model-dependent prosthesis knee controller using real-time force sensing on four terrains: rubber floor, outdoor track, grass, and sidewalk. 
(middle) Insole pressure sensor maps in the stance phases of a walking cycle.
(bottom) Measured socket and ground force profiles during stance for respective terrain.}
\label{fig:gait_tiles}
\vspace{-0.75cm}
\end{figure}

Realizing RES-CLF controllers on prostheses meets an additional challenge: 
\emph{they require knowledge of the interaction forces.} Specifically, both the ground reaction forces and moment, and socket interaction forces and moment. (For simplicity we refer to these as ``GRFs" and ``socket forces".) 
The work of \cite{gehlhar2021modeldependent} realized the first model-dependent prosthesis controller in stance with consideration for the forces---however, these forces were estimated rather than sensed.
Holonomic constraints were used to determine the GRFs, and force estimation for the socket forces.  The lack of real-time force sensing, therefore, 
necessitated the assumption of rigid contact with the ground (via the use of holonomic constraints). This inaccurately represents many real-life scenarios where the terrain deforms under a load, like granular media \cite{xiong2017stability}. Developing control methods accounting for non-rigid terrain is especially important for prostheses to enable amputees to walk stably on a variety of surfaces present in daily life. Additionally, estimating (not sensing) the socket forces may not accurately capture the varying load a user applies during stance.
To more accurately account for the interactions between the user and the prosthesis, and the prosthesis and the environment, it is necessary to integrate real-time force sensing into the model-based controller.

This work realizes the first model-dependent prosthesis knee controller that uses real-time force sensing at the ground and socket, resulting in stable human-prosthesis walking on a variety of terrains.
To achieve this result, we integrate a load cell, insole pressure sensor, and an inertial measurement unit (IMU) on a transfemoral powered prosthesis platform (shown in Fig. 1).
To leverage these sensors, we use an optimization-based controller utilizing RES-CLFs \cite{gehlhar2021separable} that directly accounts for the force sensing in real-time, i.e., the sensed forces are utilized to ``complete the model'' of the human-prosthetic system and thereby determine the next control action.
The insole pressure sensor, therefore, allows for the control actions to be dynamically modulated based upon the sensed terrain type (removing the need to assume locomotion on a non-deformable surface). The load cell quantifies the interaction between the human and prosthesis allowing the prosthesis to compensate for this dynamic load in real-time and achieve its desired behavior in the presence of these large external forces. 
This framework is demonstrated experimentally on the prosthesis shown in Fig. 1, wherein walking is achieved on 4 different terrain types with the proposed controller demonstrating improved tracking performance across all terrains.

It is important to note that force sensing has long been utilized in prosthesis control, although not in the context of realizing model-based controllers via real-time force sensing.
Load cells have been incorporated into powered prosthesis platforms to 
detect ground contact, GRFs, and center of pressure (CoP) \cite{DesignCtrlActiveProsth, VirtConsCtrlProst, elery2020design}. The work of \cite{DesignCtrlActiveProsth} used GRF sensing capability to determine motion intent to trigger transitions between gait phases of finite-state based impedance control. The work of \cite{VirtConsCtrlProst} used the CoP to encode and modulate virtual constraints for prosthesis control. However, to date, GRF and socket force measurements have not been included in the modeled dynamics to achieve model-dependent prosthesis control. Additionally, to the best of the authors' knowledge, GRFs and CoP measurements from an insole plantar pressure sensor have not been utilized as real-time feedback in prosthesis control \cite{PlantarPressureMeasurement}.

The paper is structured in the following manner. Section \ref{sec:Separable} overviews separable system control methods and RES-CLFs. Section \ref{sec:Control} constructs the controller of focus in this work that utilizes real-time force and IMU measurements. Following, Section \ref{sec:ModelGaitGen} defines the amputee-prosthesis model and describes the gait generation method. Section \ref{sec:AMPRO3} presents the pressure sensor used in this study and the powered transfemoral prosthesis platform, AMPRO3, in which the sensor is integrated. The experimental set-up and results in Section \ref{sec:Experiment} show the improved knee tracking performance with this real-time force feedback for 4 types of terrain and for 2 subjects on a single terrain type.

\section{Background on Separable Subsystems} \label{sec:Separable}
To construct the controller used as the focus of this work and justify our claim of formal guarantees of stability for the whole human-prosthesis system, we will provide a brief overview of the separable subsystem framework along with RES-CLFs starting in the context of robotic control systems.

\newsec{Robotic Control System.}
Consider a robotic control system in 2D space with $\eta$ DOFs and configuration coordinates $q = (q_l^T, q_f^T, q_s^T)^T \in \mathbb{R}^\eta$ defining the configuration space $\mathcal{Q}$. We will focus on a subsystem of the robot defined by coordinates $q_s \in \mathbb{R}^{\eta_s}$ with $m_s$ actuators. The remaining system is defined by $q_l \in \mathbb{R}^{\eta_l}$ with $m_r$ actuators. The attachment point between these systems is modeled as a 3-DOF rigid joint ($x, z$ Cartesian position, and pitch) and defined with coordinates $q_f$. Here $\eta_l + \eta_s + 3 = \eta$ and $m_s + m_r = m$, the total number of actuators. The coordinates $q_l$ include the floating base coordinates $q_B \in \R^3$. The constrained dynamics of the full system are given by the following Euler-Lagrange equation \cite{MLS}:
\begin{align} \label{eq:robotDynamics}
    & D(q) \ddot{q} + H(q, \dot{q}) = Bu + J_h^T(q) \lambda_h 
    \\ \label{eq:holoConstr}
    & J_h(q) \ddot{q} + \dot{J}_h(q, \dot{q})\dot{q} = 0,
\end{align}
where $D(q)$ is the inertia matrix, $H(q, \dot{q})$ is a vector containing Coriolis, centrifugal, and gravity forces, $B$ is the actuation matrix for $u \in \R^m$ control inputs, and $J_h(q)$ is the Jacobian of the holonomic constraints $h(q)$ and projects the constraint wrenches $\lambda_h$. 

\newsec{Robotic Subsystem.}
To develop model-based prosthesis subsystem controllers, we 
model the robotic \textit{subsystem} separately with floating base coordinates $\bar{q}_B \in \R^3$ where it attaches to the remaining system. With subsystem configuration coordinates $\bar{q} = (\bar{q}_B^T, q_s^T)^T \in \R^{\bar{\eta}}$, where $\bar{\eta} = \eta_s + 3$, the constrained robotic subsystem dynamics are:
\begin{align} \label{eq:robotSubsystem}
    & \bar{D}(\bar{q})\ddot{\bar{q}} + \bar{H}(\bar{q}, \dot{\bar{q}}) 
    = \bar{B}u_s + \bar{J}^T_h(\bar{q}) \bar{\lambda}_h + \bar{J}^T_f(\bar{q}) F_f
    \\ \label{eq:holoConstrSub}
    &\bar{J}_h(\bar{q}) \ddot{\bar{q}} + \dot{\bar{J}}_h(\bar{q}, \dot{\bar{q}}) \dot{\bar{q}} = 0.
\end{align}
Here $F_f$ are the fixed joint interaction forces inputted to this system and projected into the dynamics by $\bar{J}_f(q)$. $\bar{J}_h(\bar{q})$ is the Jacobian of the $\bar{\eta}_h$ subsystem holonomic constraints $\bar{h}(\bar{q})$ with constraint wrench $\bar{\lambda}_h$. 
By solving for $\ddot{\bar{q}}$ in the dynamics \eqref{eq:robotSubsystem} and substituting this into the holonomic constraint equation \eqref{eq:holoConstrSub}, we solve for the constraint wrenches:
\begin{equation} \label{eq:force}
\resizebox{0.89\hsize}{!}{$
    \bar{\lambda}_h = (\bar{J}_h \bar{D}^{-1} \bar{J}_h^T)^{-1} (\bar{J}_h \bar{D}^{-1}(\bar{H} - \bar{B}u_s - \bar{J}_f^T F_f) - \dot{\bar{J}}_h \dot{\bar{q}}).
    $}
\end{equation}

For the prosthesis, the work of \cite{gehlhar2021modeldependent} used \eqref{eq:force} to calculate the GRFs modeled as holonomic constraints with the ground, estimated the fixed joint force $F_f$, and calculated $\bar{q}_B$ and $\dot{\bar{q}}_B$ in stance based on inverse kinematics. For this paper we use the x-direction holonomic constraint to calculate the x-direction GRF but use a pressure sensor to measure the other GRFs: z-direction force and y-direction moment. A load cell measures the fixed joint socket forces $F_f$ and an IMU measures $\bar{q}_B$ and $\dot{\bar{q}}_B$ in non-stance.

\newsec{Separable Subsystems.}
With states $x_q = (q^T, \dot{q}^T)^T $, we write the full robotic system dynamics \eqref{eq:robotDynamics} as an ODE:
\begin{equation*}
    \dot{x}_q = 
    \underbrace{
    \begin{bmatrix}
    \dot{q} \\
    D^{-1}(q)(-H(q, \dot{q}) + J_h(q)^T \lambda_h )
    \end{bmatrix}
    }_{f_q(x_q)}
    +
    \underbrace{
    \begin{bmatrix}
    0 \\
    D^{-1}(q)(B)
    \end{bmatrix}
    }
    _{g_q(x_q)}
    u.
\end{equation*}
To apply the \textit{separable system} framework of \cite{gehlhar2019control}, we rearrange the states and define $x_r = (q_l^T, q_f^T, \dot{q}_l^T, \dot{q}_f^T)^T$ and $x_s = (q_s^T, \dot{q}_s^T)^T$ which, due to the fixed joint as explained in \cite{gehlhar2019control}, yields a system of the following form:
\begin{align} \label{eq:sepSyst}
    \begin{bmatrix}
        \dot{x}_r \\ \dot{x}_s
    \end{bmatrix}
    &=
    \underbrace{\fmatrix{}}_{f(x)}
    + \underbrace{\gmatrix{}}_{g(x)}
    \begin{bmatrix}
        u_r \\ u_s
    \end{bmatrix},
    \\ \notag
        x_r \in &\mathbb{R}^{n_r}, \,\,
        x_s \in \mathbb{R}^{n_s}, \,\,
        u_r \in \mathbb{R}^{m_r}, \,\,
        u_s \in \mathbb{R}^{m_s}.
\end{align}
The remaining control input $u_r$ does not affect the dynamics of $x_s$, allowing subsystem controller design with the $x_s$ dynamics independent of the $x_r$ dynamics. We define a \textit{separable subsystem} and \textit{remaining system} \cite{gehlhar2019control, gehlhar2021separable}, respectively,
\begin{align} \label{eq:Subsystem}
    \dot{x}_s &= f^s(x) + g^s(x) u_s, 
    \\ \label{eq:remSys}
    \dot{x}_r &= f^r(x) + g^r_1(x) u_r + g^r_2(x) u_s.
\end{align}

\newsec{Equivalent Subsystem.}
The subsystem dynamics still depend on the full system states so we alternatively express the dynamics of $x_s$ as an ODE using \eqref{eq:robotSubsystem} with a method similar to that used for the full-order dynamics:
\begin{align} \label{eq:Subsystem'}
        &\dot{\bar{x}}_s = \bar{f}^s(\mathcal{X}) + \bar{g}^s(\mathcal{X}) u_s,
        \\ \notag
        & \mathcal{X} = (\bar{x}_r^T,\, x_s^T,\, \zeta^T)^T \in \mathbb{R}^{\bar{n}}.
\end{align}
Here $\bar{x}_s = x_s$, $\bar{x}_r = (\bar{q}_B^T, \dot{\bar{q}}_B^T)^T \in \R^{\bar{n}_r}$ are \textit{measurable states}, and $\mathcal{X}$ is the vector of states $\bar{x} = (\bar{x}_r^T, x_s^T)^T$ and \textit{measurable input} $\zeta = F_f \in \mathbb{R}^{n_f}$.
This system equates to \eqref{eq:Subsystem} by a transformation $T(x) = \mathcal{X}$, where $\bar{f}^s(\mathcal{X}) = f(x)$ and $\bar{g}^s(\mathcal{X}) = g^s(x)$ for all $x$. For this robotic system form, this transformation exists and is given in \cite{gehlhar2019control}. 
An IMU and force sensor could give $\bar{x}_r$ and $\zeta$ in practice.

\newsec{Separable Subsystem RES-CLF.}
With the subsystem now defined in locally available coordinates $\mathcal{X}$, we define a class of stabilizing model-dependent subsystem controllers. Using the work of \cite{gehlhar2021separable} we construct a rapidly exponentially stabilizing control Lyapunov function (RES-CLF) $V(x_s)$ for the equivalent subsystem,
\begin{align} \label{eq:equivRESCLF}
        &c_1 \norm{x_s}^2 
        \leq 
        V(x_s) 
        \leq \frac{c_2}{\varepsilon^2} \norm{x_s}^2
        \\ \notag
        &\inf_{u_s \in \mathbb{R}^{m_s}} 
        [\dot{V}(\mathcal{X}, u_s)] 
        \leq 
        -\frac{c_3}{\varepsilon} V(x_s),
\end{align} for all $0 < \varepsilon < 1$ and $\mathcal{X} \in \R^{\bar{n}}$, with constants $c_1, c_2, c_3 > 0$. All controllers that satisfy $\dot{V}(\mathcal{X}, u_s) \leq -\frac{c_3}{\varepsilon} V(x_s)$ belong to the class $K(\mathcal{X})$:

\begin{equation} \label{eq:subK'}
    K(\mathcal{X}) = \{ u_s \in \mathbb{R}^{m_s}: 
    \dot{V} (\mathcal{X}, u_s)
    \leq
    -\frac{c_3}{\varepsilon} V(x_s) \}.
\end{equation}The work of \cite{ames2014rapidly} developed RES-CLF controllers with $\varepsilon$ to tune $\varepsilon$ to yield fast enough convergence such that a hybrid system and its zero dynamics would not be destabilized by the impact dynamics of the hybrid system. 

\newsec{Main Theoretic Idea of Previous Work.}
By the work of \cite{gehlhar2021separable}, when a RES-CLF stabilizes the remaining system \eqref{eq:remSys}, any controller $u_s \in K(\mathcal{X})$ stabilizes the full-order hybrid system with zero dynamics. Hence, separating a robotic system and constructing an equivalent subsystem provides a means to construct model-dependent subsystem controllers using solely local information guaranteeing full-order system stability while  also utilizing the natural dynamics.
In the case of a human-prosthesis system, we assume the human stabilizes itself based on research of central pattern generators which suggests biological walkers demonstrate stable rhythmic behavior \cite{Taga1995}, meaning they have limit cycles. All stabilizing controllers for these hybrid limit cycles belong to the class of RES-CLFs in \cite{gehlhar2021separable} for the remaining system.

\section{Control Methods} \label{sec:Control}
The ID-CLF-QP of \cite{reher2020inverse} provides an implementable form of a RES-CLF. We outline this construction for an equivalent robotic subsystem \eqref{eq:robotSubsystem} with the additional $\bar{J}^T_f(\bar{q}) F_f$ term in the dynamics.
We explain how we measure this term, the GRFs, and the floating base coordinates for the prosthesis dynamics.

\subsection{Controller Formulation}
To develop a Lyapunov function for the ID-CLF-QP of \cite{reher2020inverse} we construct outputs to feedback linearize the system and dictate the motion of the robot.

\newsec{CLF Construction.}
We define linearly independent subsystem positional outputs to enforce desired trajectories on the robotic subsystem, 
\begin{equation} \label{eq:subOutputs}
    y_s(x_s) = y^a_s(x_s) - y_s^d(\tau(x_s), \alpha),
\end{equation}
where $y^a_s(x_s)$ are the actual joint outputs and $y_s^d(\tau(x_s, \alpha))$ are the desired trajectories defined by parameters $\alpha$ and modulated by $\tau(x_s)$, a state-based phase variable. 
Taking the derivatives along the equivalent subsystem dynamics \eqref{eq:Subsystem'}, we relate the outputs to the control input $u_s$:
\begin{equation*}
    \ddot{y}_s= L_{\bar{f}^s}^{2}y_s(\mathcal{X}) + L_{\bar{g}^s}L_{\bar{f}^s} y_s(\mathcal{X}) u_s.
\end{equation*}
Here $L_{\bar{f}^s}^{2}y_s(\mathcal{X})$ and $L_{\bar{g}^s}L_{\bar{f}^s} y_s(\mathcal{X})$ denote the Lie derivatives \cite{IsidoriNonlinSyst}. Because the outputs are linearly independent, $L_{\bar{g}^s}L_{\bar{f}^s} y_s(\mathcal{X})$ is invertible, making our system feedback linearizable \cite{IsidoriNonlinSyst} with feedback linearizing controller, 
\begin{equation} \label{eq:feedlin}
    u_s(\mathcal{X}) = \big(L_{\bar{g}^s}L_{\bar{f}^s} y_s(\mathcal{X}) \big)^{-1} \big( -L_{\bar{f}^s}^{2}y_s(\mathcal{X}) + \nu \big),
\end{equation}
with auxiliary control input $\nu$. This controller yields $\ddot{y}_s = \nu$ and the following linearized output dynamics with linear system coordinates $\xi = (y_s^T, \dot{y}_s^T)^T$,
\begin{equation*}
    \dot{\xi} = 
    \underbrace{
    \begin{bmatrix}
    0 & I \\
    0 & 0
    \end{bmatrix}
    }_F
    \xi
    +
    \underbrace{
    \begin{bmatrix}
    0 \\
    I
    \end{bmatrix}
    }_G
    \nu.
\end{equation*}
We solve the
continuous time algebraic Riccati equation,
\begin{equation*}
    F^T P + PF - PGG^T P + Q = 0,
\end{equation*}
for this linear system with user-selected weighting matrix $Q = Q^T >0$,
for $P = P^T > 0$ to construct a RES-CLF per the methods of \cite{ames2014rapidly}:
\begin{equation*} \label{eq:subRES-CLF}
    V(\xi) = \xi^T 
    \begin{bmatrix}
    \frac{1}{\varepsilon} I & 0 \\
    0 & I
    \end{bmatrix}
    P
    \begin{bmatrix}
    \frac{1}{\varepsilon} I & 0 \\
    0 & I
    \end{bmatrix}
    \xi
    =: \xi^T P^\varepsilon \xi.
\end{equation*}
Taking the derivative gives the convergence constraint:
\begin{equation*}
    \dot{V}(\xi, \nu) = L_F V(\xi) + L_G V(\xi) \nu \leq - \frac{1}{\varepsilon} \frac{\lambda_{min} (Q)}{\lambda_{max} (P)} V(\xi),
\end{equation*}
with Lie derivatives along the linearized output dynamics as,
\begin{align*}
L_F V(\xi) &= \xi^T (F^T P_\varepsilon + P_\varepsilon F) \xi,
\\ 
L_G V(\xi) &= 2 \xi^T P_\varepsilon G.
\end{align*}

\newsec{ID-CLF-QP.}
We can write this RES-CLF and its derivative in terms of $x$ and $\mathcal{X}$ to use in the ID-CLF-QP since $\xi$ depends on $x_s$ through $y_s(x_s)$ and $\dot{y}_s(x_s)$ and the relationship in \eqref{eq:feedlin} shows $\nu$ depends on $\mathcal{X}$:
\begin{equation} \label{eq:nu}
    \nu =  L_{\bar{f}^s}^{2}y_s(\mathcal{X}) + L_{\bar{g}^s}L_{\bar{f}^s} y_s(\mathcal{X})    u_s(\mathcal{X}).
\end{equation}
We hence obtain the subsystem RES-CLF \eqref{eq:equivRESCLF} with
$c_1 = \lambda_{min}(P)$, $c_2 = \lambda_{max}(P)$, and $c_3 = \frac{\lambda_{min} (Q)}{\lambda_{max} (P)}$.

To avoid the matrix inversions required by \eqref{eq:nu}, we use the facts that $\nu = \ddot{y}_s$ and the output $y_s(x_s)$ is a positional constraint, i.e. $y_s(\bar{q})$, to write the $\ddot{y}_s$ in terms of the robotic subsystem's configuration coordinates $\bar{q}$:
\begin{equation} \label{eq:yddot}
    \ddot{y}_s = \underbrace{\frac{\partial}{\partial \bar{q}} \bigg( \frac{\partial y_s}{\partial \bar{q}} \dot{\bar{q}} \bigg)}_{\dot{J}_y(\bar{q}, \dot{\bar{q}})} \dot{\bar{q}} 
    + 
    \underbrace{ \frac{\partial y_s}{\partial \bar{q}}}_{J_y(\bar{q})} \ddot{\bar{q}}.
\end{equation}
This formulation equates to \eqref{eq:nu}, as shown by \cite{reher2020inverse}. 

To prescribe a PD controller to $\ddot{y}$ with a control input $u_s$ close to \eqref{eq:feedlin}, we define $\nu = K_p y^s(x_s) + K_d \dot{y}^s(x_s) := \nu_{\rm{pd}}$ and minimize the difference between $\nu_{\rm{pd}}$ and \eqref{eq:yddot} in the QP cost. We also include the holonomic constraints in the cost as soft constraints since these are difficult to satisfy precisely on hardware. With decision variables $\Upsilon = (\ddot{\bar{q}}^T, u_s^T, \bar{\lambda}_h^T, \delta)^T \in \mathbb{R}^{\eta_v}$, with $\eta_v = \bar{\eta} + m_s + \bar{\eta}_h + 1$, and using the terms, 
\begin{align*}
    &J_c(\bar{q}) = 
    \begin{bmatrix}
    J_y(\bar{q}) \\
    \bar{J}_h(\bar{q})
    \end{bmatrix},
    &\dot{J}_c(\bar{q}, \dot{\bar{q}}) =
    \begin{bmatrix}
    \dot{J}_y(\bar{q}, \dot{\bar{q}}) \\
    \dot{\bar{J}}_h(\bar{q}, \dot{\bar{q}})
    \end{bmatrix},
\end{align*}
we formulate our \textbf{ID-CLF-QP}:
\begin{equation} \label{eq:ID-CLF-QP+Ff}
\resizebox{0.89\hsize}{!}{$
\begin{aligned}
\Upsilon^\star 
= \mathop {\argmin }\limits_{{\Upsilon \in \mathbb{R}^{\eta_v}}} \,
& \Big|\Big| \dot{J}_c(\bar{q}, \dot{\bar{q}}) \dot{\bar{q}} + J_c(\bar{q}) \ddot{\bar{q}} - \mu^{\rm{pd}} \Big|\Big|^2 
+ \sigma W(\Upsilon) 
+ \rho \delta
 \\ 
\textrm{s.t.} \,\, 
& \bar{D}(\bar{q})\ddot{\bar{q}} + \bar{H}(\bar{q}, \dot{\bar{q}}) 
= \bar{B}u_s + \bar{J}^T_h(\bar{q}) \bar{\lambda}_h + \bar{J}^T_f(\bar{q}) F_f
\\ 
& L_FV(\mathcal{X}) + L_GV(\mathcal{X}) \big(\dot{J}_y \dot{\bar{q}} + J_y \ddot{\bar{q}}\big) \leq - \frac{\gamma}{\varepsilon} V(\mathcal{X}) + \delta
\\
& -u_{\rm{max}} \leq u_s \leq u_{\rm{max}}. 
\end{aligned}
$}
\end{equation}
Here $\mu^{\rm{pd}} = (\nu_{\rm{pd}}^T, 0^T )^T$, the regularization term $W(\Upsilon)$ makes the system well-posed, $\sigma$ and $\rho$ are user-selected weights, and the relaxation term $\delta$ ensures the torque bounds $(-u_{\rm{max}}, u_{\rm{max}})$ are always feasible. (We leave out the arguments on $J_y, \dot{J}_y$ for notational simplicity.)

\subsection{Controller Realization for Hardware}

While \cite{gehlhar2021modeldependent} realized a variation of the ID-CLF-QP, this method was only applied in stance, relied on a force estimation method for $F_f$, and used the holonomic constraint wrench $\bar{\lambda}_h$ for the GRFs. This rigid-contact model used for the GRFs does not hold for a foot contacting a variety of real-world non-rigid terrains. To overcome these limitations, this paper incorporated an IMU, load cell, and insole pressure sensor into the prosthesis platform.

\newsec{Sensor Measurements to Complete Dynamics.}
While the floating base positions and velocities can still be obtained in stance by inverse kinematics with the prosthesis joint positions and velocities, we use an IMU on the human leg adapter to measure the floating base y-rotation and angular velocity in swing. These measurements with the kinematics give the x and z-Cartesian velocities. The x and z-Cartesian positions do not affect the dynamics and hence are not required.
This paper employed a 6-axis load cell to directly measure the interaction forces $F_f$ between the human and prosthesis and an insole pressure sensor, detailed in Section \ref{sec:AMPRO3}, to determine the vertical GRF $F_{g, z}$ and pitch ground reaction moment $M_{g, y}$.
The only remaining unknown force is the horizontal GRF. We solve for the wrench $\lambda_{h,x} \in \R^1$ through a holonomic constraint, assuming the foot does not slip on the ground.

\newsec{Force Sensing ID-CLF-QP.}
The final controller formulation is
\begin{equation} \label{eq:ID-CLF-QP+Ff+Fg}
\resizebox{0.89\hsize}{!}{$
\begin{aligned}
\Upsilon^\star 
= \mathop {\argmin }\limits_{{\Upsilon \in \mathbb{R}^{\eta_v}}} \,
& \Big|\Big| \dot{J}_c(\bar{q}, \dot{\bar{q}}) \dot{\bar{q}} + J_c(\bar{q}) \ddot{\bar{q}} - \mu^{\rm{pd}} \Big|\Big|^2 
+ \sigma W(\Upsilon) 
+ \rho \delta
 \\ 
\textrm{s.t.} \,\, 
& \bar{D}(\bar{q})\ddot{\bar{q}} + \bar{H}(\bar{q}, \dot{\bar{q}}) 
= \bar{B}u_s + \bar{J}^T_h(\bar{q}) \tilde{F}_g + \bar{J}^T_f(\bar{q}) F_f
\\ 
& L_FV(\mathcal{X}) + L_GV(\mathcal{X}) \big(\dot{J}_y \dot{\bar{q}} + J_y \ddot{\bar{q}}\big) \leq - \frac{\gamma}{\varepsilon} V(\mathcal{X}) + \delta
\\
& -u_{\rm{max}} \leq u_s \leq u_{\rm{max}}, 
\end{aligned}
$}
\end{equation}
with modified set of decision variables $\tilde{\Upsilon} = (\ddot{\bar{q}}^T, u_{s}^T, \bar{\lambda}_{h, x}, \delta)^T \in \mathbb{R}^{\tilde{\eta}_v}$ and $\tilde{\eta}_v = \bar{\eta} + m_s + 2$. The decision variable $\bar{\lambda}_{h, x}$ is included with the measured GRFs $F_{g, z}$ and $M_{g, y}$ in $\tilde{F}_{g} = (\bar{\lambda}_{h, x}, F_{g, z}, M_{g, y})^T$. 

\section{Amputee-Prosthesis Model and Gait Generation} \label{sec:ModelGaitGen}
To develop outputs for the human-prosthesis system for the ID-CLF-QP, we construct a model of the system and use hybrid zero dynamics trajectory generation methods.

\begin{figure} [t]  
\centering
\includegraphics[width=1\columnwidth]{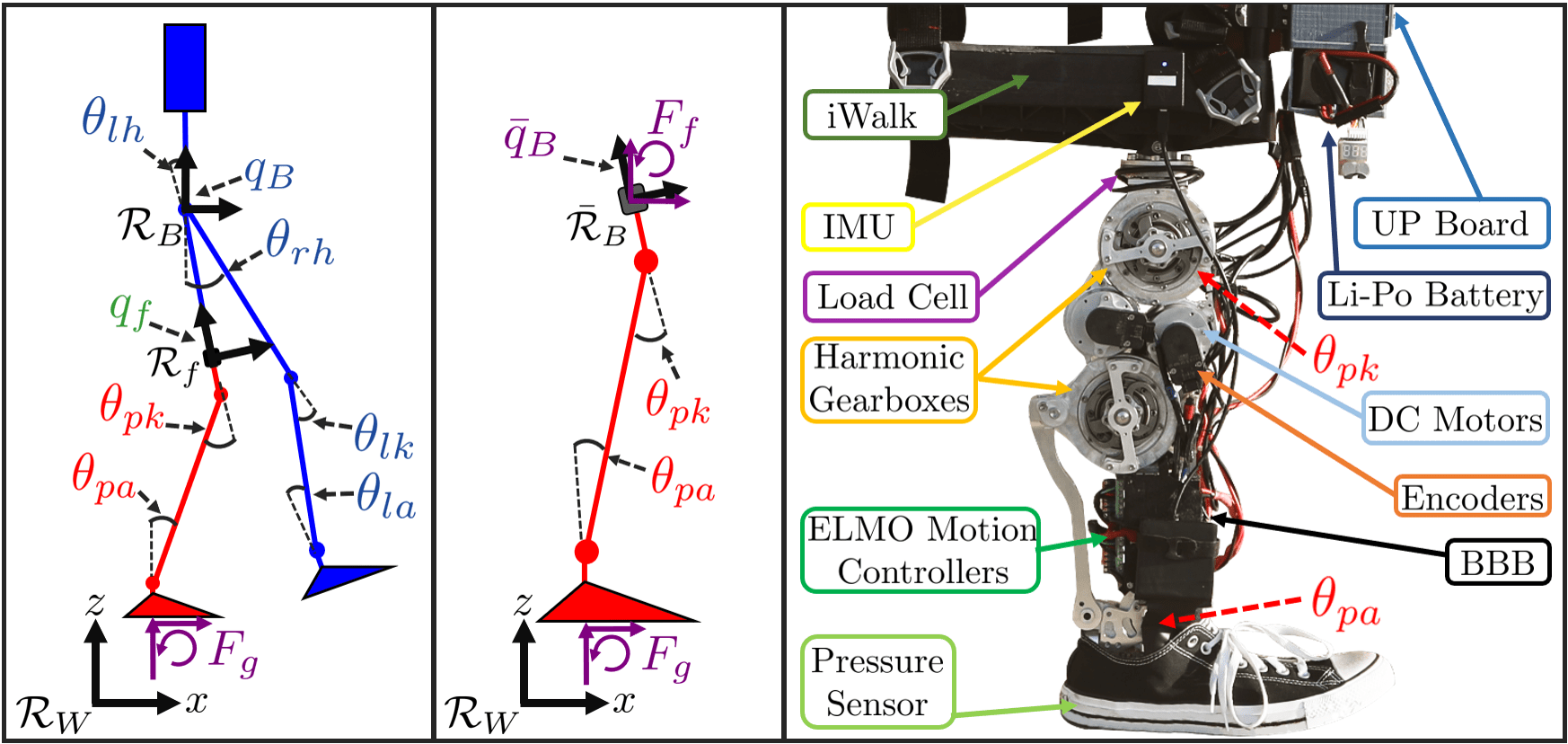}
\vspace{-0.8cm}
\caption{(Left) The separable human-prosthesis system with full system configuration coordinates. (Middle) The prosthesis equivalent subsystem with subsystem configuration coordinates. (Right) Transfemoral powered prosthesis AMPRO3 with labeled hardware components.}
\label{fig:ampro3}
\vspace{-0.7cm}
\end{figure}

\newsec{Amputee-Prosthesis Model.} 
We model an amputee-prosthesis system as a bipedal robot with 8 links and 12 DOFs, i.e. $\eta = 12$ for \eqref{eq:robotDynamics}. The prosthesis subsystem has coordinates $q_s = (\theta_{pk}$, $\theta_{pa})^T$ for the knee and ankle pitch respectively, making $\eta_s = 2$. The remaining amputee system has coordinates $q_l = (q_B^T, \theta_{lh}, \theta_{lk}, \theta_{la}, \theta_{rh})^T$ defining the floating base at the torso and the left hip, left knee, left ankle, and left hip pitch joints respectively. The fixed joint coordinates $q_f$ define the interface between the amputee's partial thigh and the top of the prosthesis. See Fig. \ref{fig:ampro3}. Every non-fixed joint is actuated, giving $m_s = 2$ and $m_r = 4$.

The human subject's height and weight along with human inertia, limb mass, and limb length percentage data from \cite{HumanParam, HumanInertia} provide the human parameters. 
We base the prosthesis model parameters on the CAD model of AMPRO3 \cite{zhao2017preliminary}, our transfemoral powered prosthesis.
This full system model gives the dynamics of \eqref{eq:robotDynamics}. The prosthesis equivalent subsystem with floating base coordinates $\bar{q}_B$ at the socket gives the dynamics \eqref{eq:robotSubsystem} used in 
\eqref{eq:ID-CLF-QP+Ff+Fg}.

For this initial realization of a force sensing model-dependent prosthesis controller, we limit the scope to a knee controller. We use the given model in the ID-CLF-QP to generate a knee torque, but predefine a torque for the ankle based on a varying set point PD controller. Since we do not enforce an ankle trajectory, we generate a prosthesis knee trajectory for the prosthesis using the full system model without ankles. Future work will realize this force sensing ID-CLF-QP controller on both the knee and ankle for a more complex gait that emulates human heel-toe roll.

\newsec{Hybrid Systems and Human-Like Gait Generation.}
Because bipedal walking contains both discrete and continuous dynamics, we model the amputee-prosthesis system as a hybrid system \cite{ames2014human}. Since the amputee-prosthesis system is asymmetrical, we consider two domains of continuous dynamics \eqref{eq:robotDynamics} $\Domain_v$ with indices $v \in \{\rm{ps}, \rm{pns}\}$ for prosthesis stance and for prosthesis non-stance, respectively. Each domain has a 3 dimensional holonomic constraint for the respective stance foot ground contact in addition to the fixed joint constraint. Events connect these domains together in a directed cycle, specifically the event of the non-stance foot contacting the ground. The work of \cite{ModelsGrizzle} explains the impact dynamics occurring at foot-strike.

To design output trajectories for the amputee-prosthesis system that are invariant through impact, we use a hybrid zero dynamics condition \cite{westervelt2018feedback} in an optimization whose solution must also satisfy the dynamics and feasibility constraints. We design the cost function to minimize the difference between the outputs (the joints) and human joint kinematic walking data obtained through motion capture \cite{gehlhar2020data}. The optimization yields parameters for each domain $\alpha_v$ that define \Bezier polynomials for the desired trajectories $y^d_{s, v}(\tau(x_s), \alpha_v)$ parameterized by the state-based phase variable $\tau(x_s)$, forward hip position, which goes from 0 to 1 in each domain $\Domain_v$ \cite{westervelt2018feedback}. Details given \cite{gehlhar2020data}. The resulting trajectories match the human data well and are shown in \cite{gehlhar2021modeldependent}.
Since the optimization gives a prosthesis trajectory that matches human walking data and is stable with the human side emulating the human data, we assume in practice the human will still be able to stabilize itself with the prosthesis following this human-like trajectory. This satisfies the required condition about the human for our main theoretical idea in \ref{sec:Separable} to ensure full human-prosthesis system stability for a RES-CLF controlling the prosthesis.

\section{Prosthesis Platform for Controller Realization} \label{sec:AMPRO3}
We integrated a pressure sensor, a load cell, and IMU into a current powered prosthesis platform to achieve model-based prosthesis control with real-time in-the-loop forcing sensing.
First, we will describe the pressure sensor selection and use due to the novelty of incorporating plantar pressure data as real-time force input to a model-dependent controller.

\newsec{Pressure Sensor.}
The pressure sensor used for this study met the restrictive real-time control requirements for our application. Most commercially available insole pressure sensors are designed for recording data for offline gait analysis and
are incompatible with our application. The sensor for this study is \$2.5-3k and a company-provided API returns the raw data in real-time over a USB connection.

The pressure sensor used is a SensorProd Inc. Tactilus Foot Insole Sensor System, High-Performance V Series (SP049).
Made of a piezoresistive sensor array, the insole pressure sensor can sense up to 206.8 kPa at 101 separate points per foot. 
The sensor provides a resolution of $\pm$1 $\mu$Pa, along with an accuracy of $\pm$10\%, repeatability of $\pm$2\%, and a hysteresis of $\pm$5\%.
To interface with the pressure sensor, we use an UP Board (02/32), a small x86 single-board computer.

We incorporated the Tactilus API, a precompiled C++ Windows Library from SensorProd Inc., into a Windows C++ program which scanned the pressure readings in real time at about 200 Hz with the UP Board through a USB connection. 
We applied a Gaussian smoothing filter and simple moving average filter to the sensor element pressure readings. With the sensor element pressure, surface area, and displacement from the ankle's center of rotation we calculate the vertical GRF $F_{g, z}$ and ground reaction moment $M_{g, y}$.

\newsec{Prosthesis Platform AMPRO3.}
We use the transfemoral powered prosthesis platform AMPRO3 custom-built and introduced in \cite{zhao2017preliminary}. Two brushless DC motors (MOOG BN23) with 1 Nm peak torque actuate the knee and ankle pitch joints through interactions with their respective timing belt connected to each joint's harmonic gear box. This gear reduction system gives a 120:1 mechanical reduction for the knee and 175:1 for the ankle. 
The motors are controlled by 2 ELMO motion controllers (Gold Solo Whistle) which receive position and velocity feedback from 2 incremental encoders. These motion controllers in turn send this feedback to the microprocessor. The microprocessor returns a commanded torque to the motion controllers. The controller algorithms run on the Beaglebone Black Rev C (BBB) microprocessor at 166Hz and are coded in C++ packages with ROS. The coded force sensing ID-CLF-QP is based on code from \cite{reher2020inverse}. 

The insole sensor is physically integrated into the prosthetic system through placement over the insole of a shoe worn by the prosthesis foot. The sensor connects to the UP Board through USB, and sends force and moment measurements to the BBB through UDP over Ethernet. There is a 5 ms time delay between when the sensor is read and the BBB receives the data caused largely by the time it takes the Windows program to receive all the data from the sensor.

\begin{figure}
\centering
  \includegraphics[width=0.9\columnwidth]{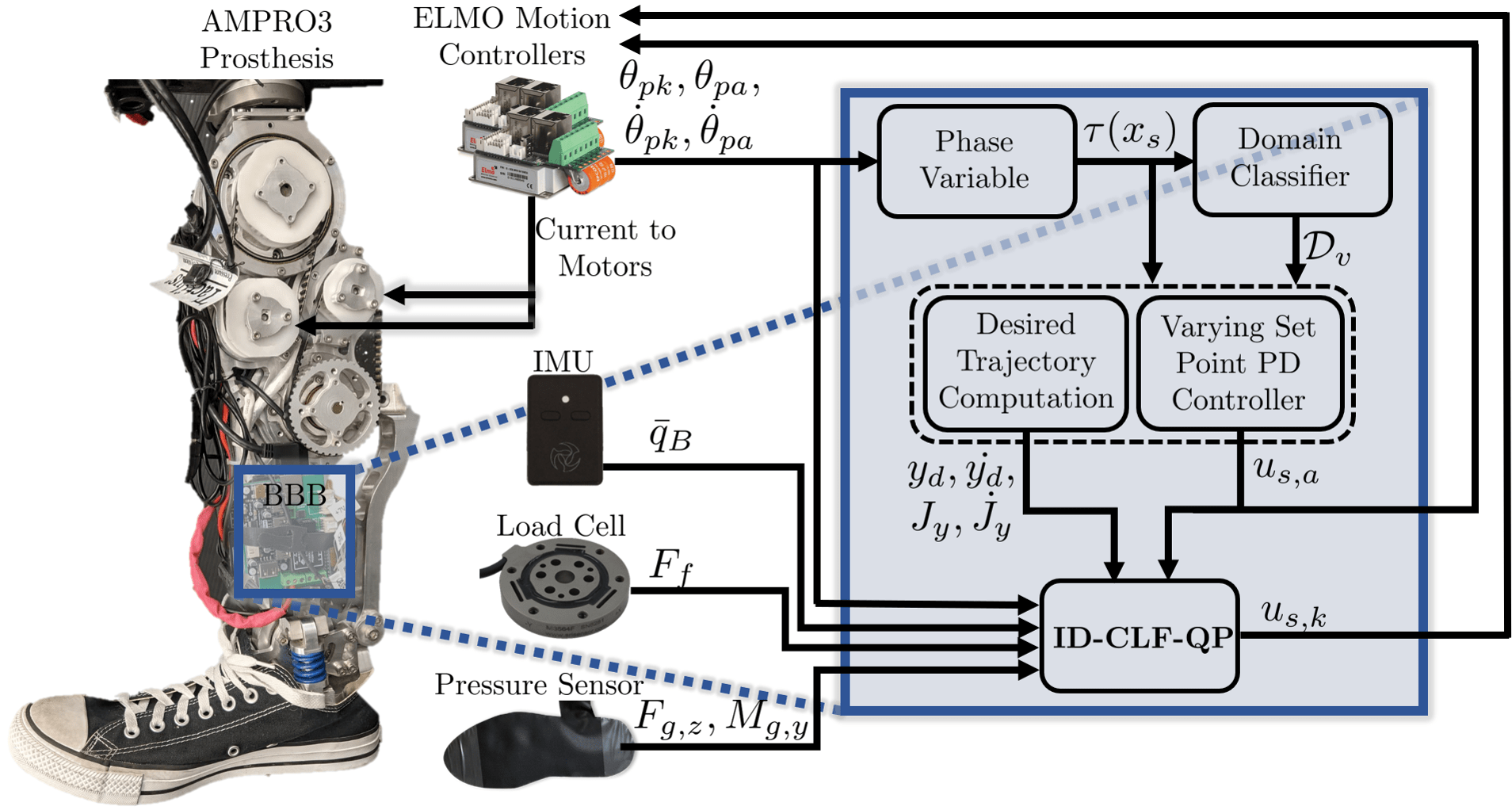}
  \vspace{-0.4cm}
  \caption{Block diagram depicting how the Beaglebone microprocessor, motion controllers, IMU, load cell, and pressure sensor are involved in the control scheme.}
  \vspace{-0.6cm}
   \label{fig:flowdiagram}
\end{figure}

A Yost Labs 3-Space™ Sensor USB/RS232 IMU is mounted to the side of the human leg adapter on the prosthesis platform. This connects to the BBB via USB and streams data at 750 Hz.
A 6-axis load cell (M3564F, Sunrise Instruments) is mounted at the interface between the proximal end of the prosthesis knee joint and distal end of the human leg adapter with custom designed aluminum parts.
The load cell can measure forces/moments along the x and y-axes up to 2500 N/200 Nm and z-axis up to 5000 N/100 Nm. A signal conditioning box (M8131 Sunrise Instruments) connects to the BBB with a custom-designed cape and sends data and sends new data via CAN communication once every control loop. 
A newly designed electronics box mounted on the iWalk houses the load cell box, UP Board, and a 9-cell 4400 mAh Li-Po battery (Thunder Power RC) which powers the whole system.
Fig. \ref{fig:ampro3} shows AMPRO3 with the aforementioned components labeled and Fig. \ref{fig:flowdiagram} shows a control block diagram with the sensors.
The prosthesis weighs 5.95 kg on its own, and totals 10.54 kg with the iWalk, electronics box, battery, and sensors.

\section{Human-Prosthesis Experimentation} \label{sec:Experiment}
On this prosthesis platform, we realize our model-dependent force sensing knee controller, resulting in stable human-prosthesis walking. We present the results here.

\begin{figure}
\centering
  \includegraphics[width=1\columnwidth]{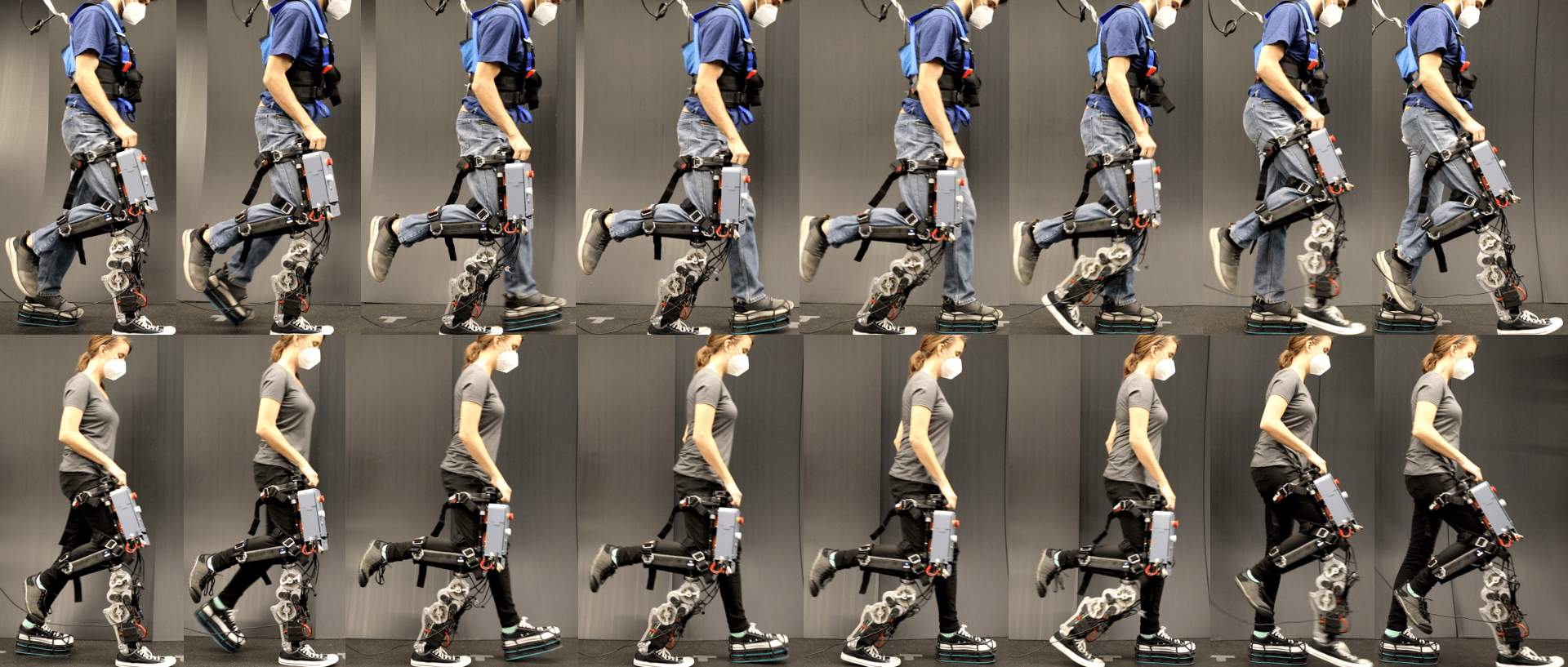}
  \vspace{-0.6cm}
  \caption{Gait tiles of 2 subjects walking on a rubber floor with the force sensing ID-CLF-QP prosthesis controller.}
  \vspace{-0cm}
   \label{fig:gait_tiles_all}
\end{figure}

\newsec{Experimental Procedure.}
A 1.7 m, 62 kg non-amputee human subject (Subject 1) and a 1.8 m, 75 kg non-amputee subject (Subject 2) tested the prosthesis device with an iWalk adapter. The iWalk allows a subject's bent right leg to be strapped to the device for walking as shown in Fig. \ref{fig:gait_tiles}. A foam shoe lift strapped to the bottom of their left leg's shoe evens the length difference between their own left leg and their right leg with the prosthesis. We applied the proposed controller to the knee in both stance and non-stance phase.
A PD controller with varying set-point was applied to the prosthesis ankle.
Both subjects walked with the prosthesis for at least 30 steps with 4 different controllers on a rubber floor. Gait tiles of an experiment are shown in Fig. \ref{fig:gait_tiles_all}. Subject 1 walked with the prosthesis with 2 of these controllers on an additional 3 terrains: an outdoor track, grass, and a sidewalk.

The first controller is the force sensing ID-CLF-QP \eqref{eq:ID-CLF-QP+Ff+Fg}, the controller of focus in this paper, using the pressure sensor to obtain $F_{g, z}$, $M_{g, y}$, the load cell to obtain $F_f$, and the IMU in swing for $\bar{q}_B$ and $\dot{\bar{q}}_B$. 
For comparison, we tested an ID-CLF-QP without including the measurements from the pressure sensor and load cell to see the effect the force sensor measurements have on the controller performance. Here the GRFs were determined with the holonomic constraints \eqref{eq:holoConstrSub} but the fixed joint forces $F_f$ were considered 0. 
Thirdly, we compared the performance to the previous force estimating ID-CLF-QP in \cite{gehlhar2021modeldependent}, where the holonomic constraints are again used to determine the GRFs and the effect of the socket forces is estimated with a force estimator with a moving average time window of 30. Here the time window was increased to yield a smooth torque response when using the same gains  $K_P, K_D$ in $\nu_{\rm{pd}}$ as those selected for the force sensing ID-CLF-QP. 
The gains $\nu_{\rm{pd}}$ and all other user-selected terms in the QP were kept consistent between controllers.
Finally, we compared the performance to a traditional PD controller, which is used in other prosthesis control methods \cite{VirtConsCtrlProst,gregg2021nonholonomic}.
The experimental results are shown in the supplemental video \cite{youtube}.

\begin{figure} [t]  
\centering
\includegraphics[width=1\columnwidth]{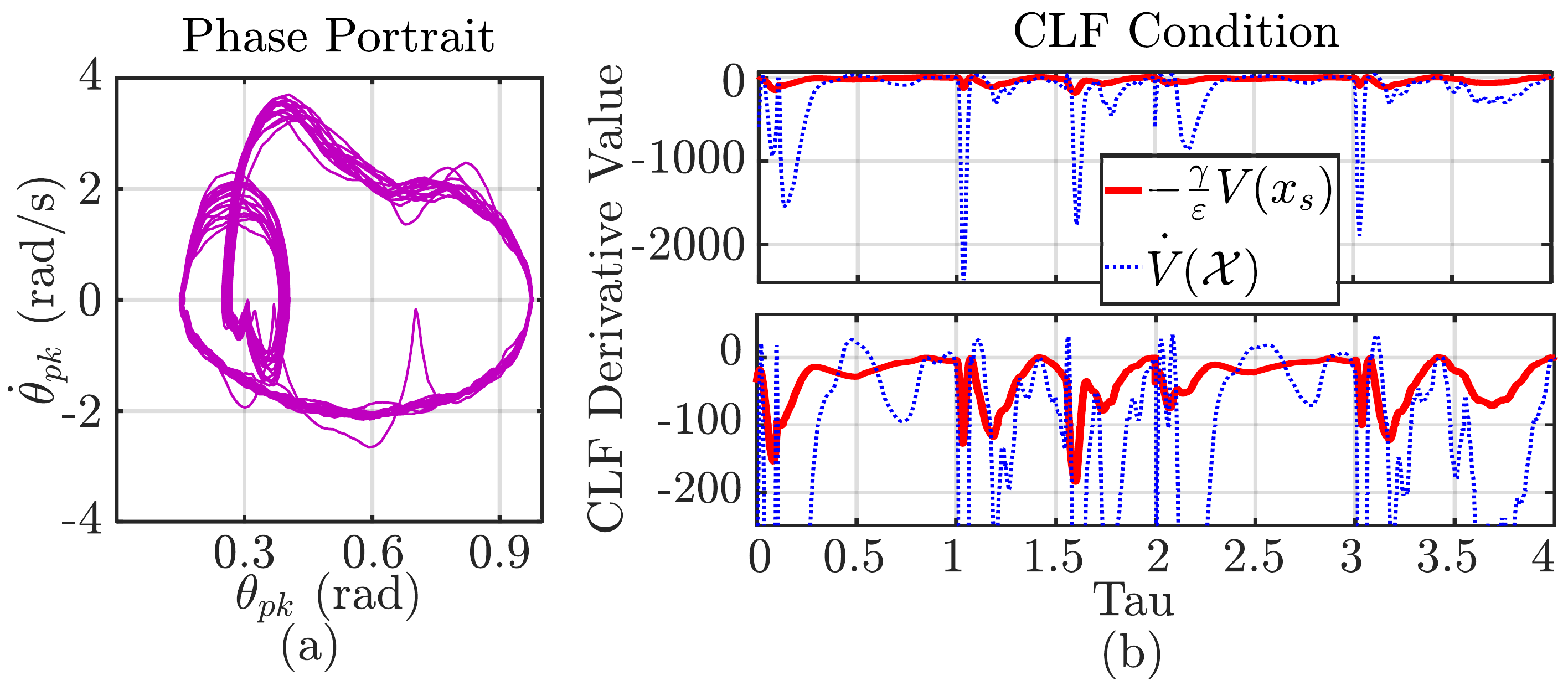}
\vspace{-0.6cm}
\caption{(a) Phase portrait of first 18 steps using the force sensing ID-CLF-QP controller \eqref{eq:ID-CLF-QP+Ff+Fg} on a sidewalk, showing the system yields a stable periodic orbit. (b) (top) CLF bound and derivative for first 4 steps of the same experiment, showing the prosthesis usually satisfies the stability condition, (bottom) magnified plot of the CLF bound, w.r.t. phase variable $\tau(x_s)$.}
\label{fig:clf_control_input}
\vspace{-0.6cm}
\end{figure}  

\begin{figure} [t]  
\centering
\includegraphics[width=1\columnwidth]{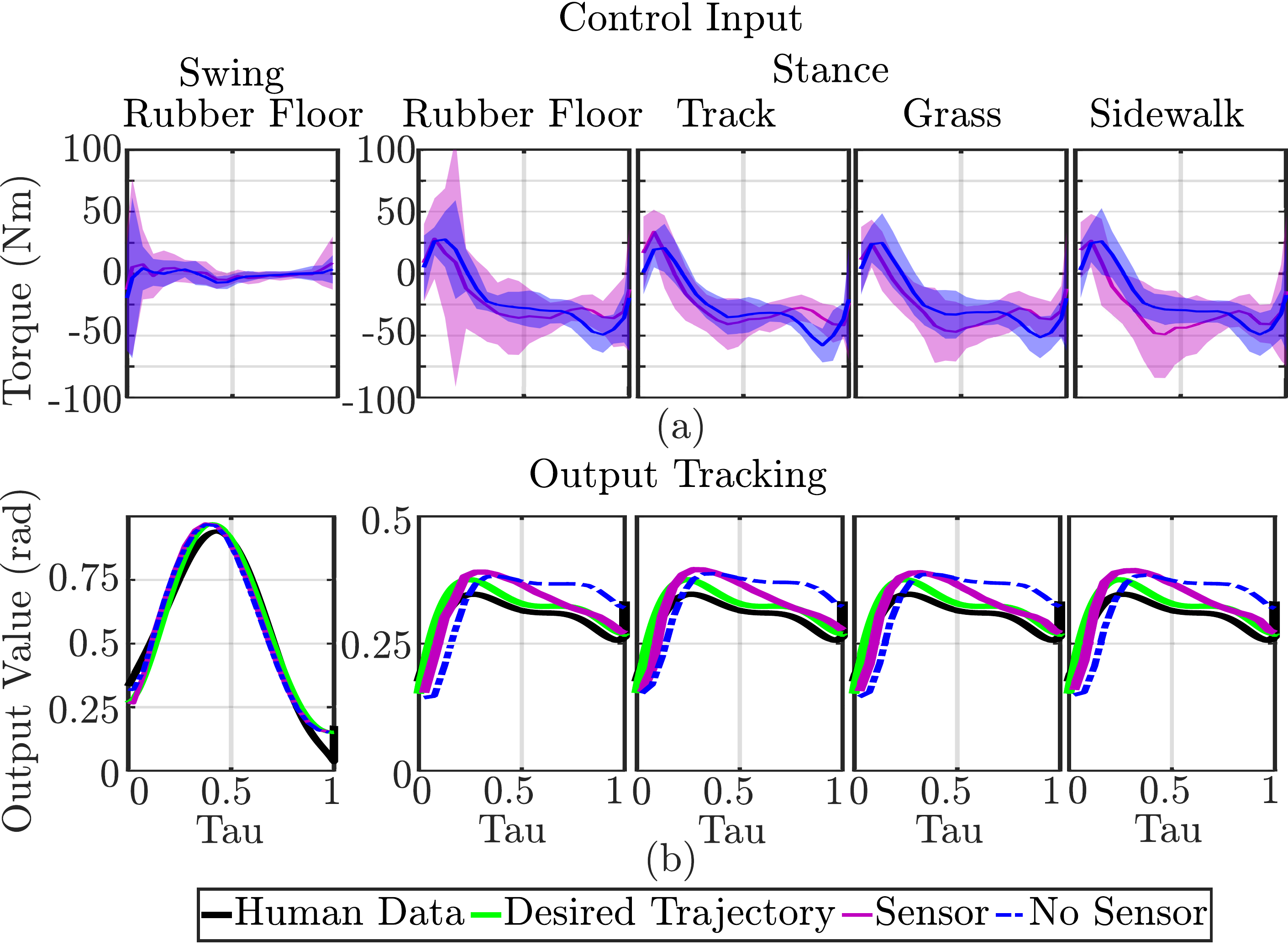}
\vspace{-0.25cm}
\caption{(a) Control input (mean and 3 standard deviations) and (b) output tracking with force sensing ID-CLF-QP and the ID-CLF-QP without force sensors for four different terrains for 11 step cycles, w.r.t. to phase variable $\tau(x_s)$.}
\label{fig:outputs}
\vspace{-0.4cm}
\end{figure}    

\begin{figure} [t]  
\centering
\includegraphics[width=1\columnwidth]{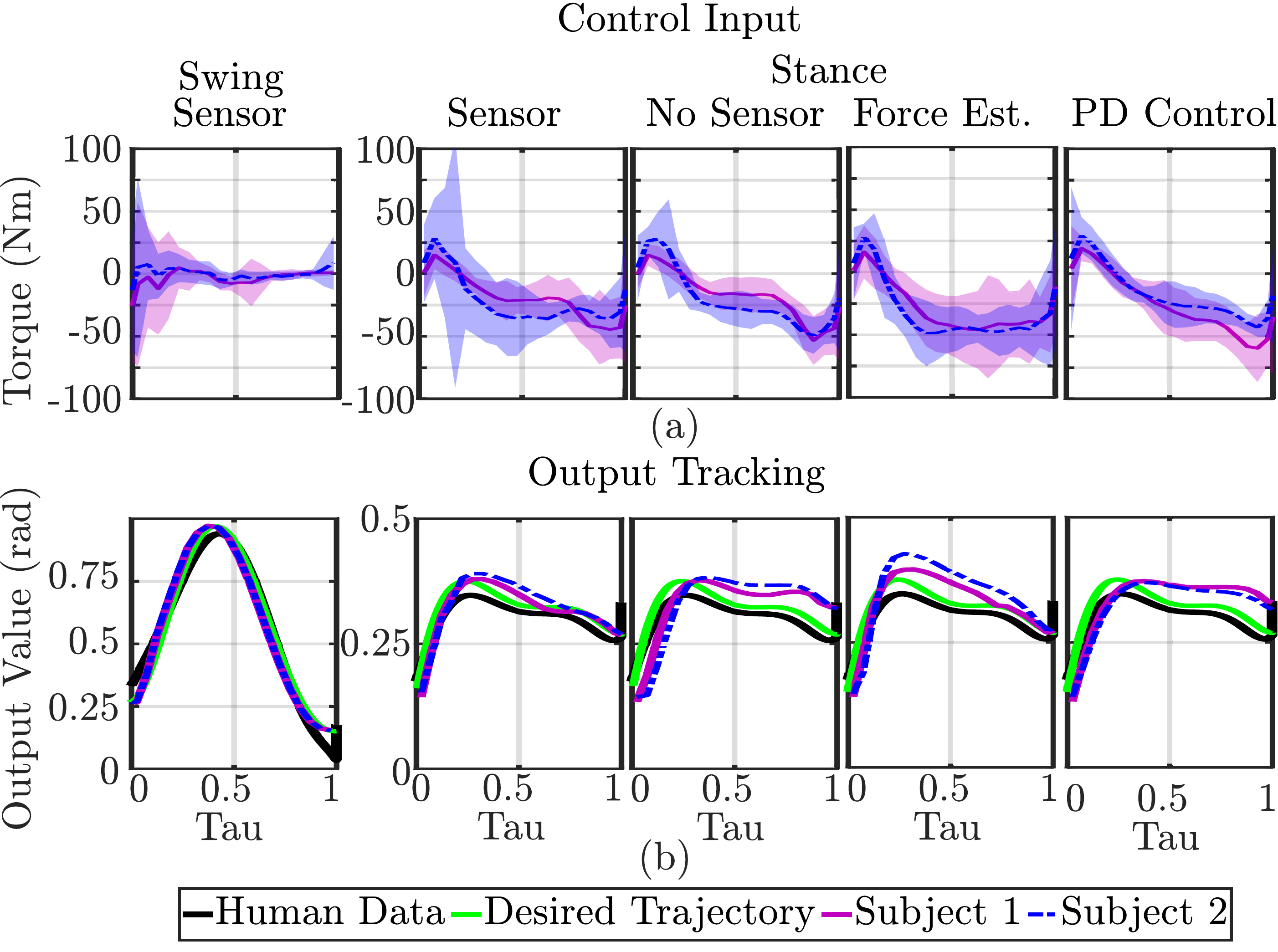}
\vspace{-0.25cm}
\caption{(a) Control input (mean and 3 standard deviations) and (b) output tracking (mean) with force sensing ID-CLF-QP, no sensor ID-CLF-QP, force estimating ID-CLF-QP, and PD controller for 11 step cycles, w.r.t. phase variable $\tau(x_s)$.}
\label{fig:subjects}
\vspace{-0.6cm}
\end{figure}

\begin{table}[t]
\caption{Tracking RMSE of 4 Terrains for 2 Controllers\label{fixedweights}}
\vspace{-0.2cm}
\setlength{\tabcolsep}{7pt}
\begin{tabular}{l|ll|ll|}
\cline{2-5}
                                 & \multicolumn{2}{l|}{Stance RMSE ($\theta$)} & \multicolumn{2}{l|}{Swing RMSE ($\theta$)} \\ \cline{2-5} 
                                    & Sensor           & No Sensor        & Sensor           & No Sensor       \\ \hline
\multicolumn{1}{|l|}{Rubber Fl.}    & 0.0409           & 0.0797           & 0.0341           & 0.0349          \\
\multicolumn{1}{|l|}{Track}         & 0.0460           & 0.0662           & 0.0302           & 0.0297          \\
\multicolumn{1}{|l|}{Grass}         & 0.0398           & 0.0740           & 0.0328           & 0.0307          \\
\multicolumn{1}{|l|}{Sidewalk}      & 0.0380           & 0.0763           & 0.0331           & 0.0295          \\ \hline 
\end{tabular}%
\label{tab:terrains}

\vspace{0.25cm}

\caption{Tracking RMSE of 4 Controllers for 2 Subjects\label{fixedweights}}
\vspace{-0.2cm}
\setlength{\tabcolsep}{6pt}
\begin{tabular}{l|ll|ll|}
\cline{2-5}
                                 & \multicolumn{2}{l|}{Stance RMSE ($\theta$)} & \multicolumn{2}{l|}{Swing RMSE ($\theta$)} \\ \cline{2-5} 
                                 & Subject 1        & Subject 2        & Subject 1        & Subject 2       \\ \hline
\multicolumn{1}{|l|}{Sensor}     & 0.0237           & 0.0409           & 0.0331           & 0.0341          \\
\multicolumn{1}{|l|}{No Sensor}  & 0.0455           & 0.0797           & 0.0372           & 0.0349          \\
\multicolumn{1}{|l|}{Force Est}  & 0.0357           & 0.0552           & 0.0409           & 0.0401          \\
\multicolumn{1}{|l|}{PD Control} & 0.0484           & 0.0522           & 0.0464           & 0.0427          \\ \hline 
\end{tabular}%
\label{tab:subjects}

\vspace{-0.4cm}
\end{table}

\newsec{Hardware Results.}
The knee phase portrait of Fig. \ref{fig:clf_control_input} a. shows the stability of this main controller for 18 steps on a sidewalk.
Fig. \ref{fig:clf_control_input} b. depicts the CLF derivative with its upper CLF stability bound for the first 4 steps. While the derivative occasionally exceeds the bound because of the relaxation term in the QP \eqref{eq:ID-CLF-QP+Ff+Fg}, the overall human-prosthesis system can still be input-to-state stable as proved in \cite{gehlhar2021estimate}. Additionally, although the force measurements may have error from sensor noise and time delay, the work of \cite{gehlhar2021estimate} showed for a bounded error in the force measurement the prosthesis will still be stable to a set and conditions exist such that the human will remain exponential input-to-state stable \cite{SontagCharacISS} when the prosthesis deviates from its nominal control law.

Fig. \ref{fig:outputs} a. shows the mean torque inputs with 3 standard deviations of the first two controllers for 11 step cycles on 4 different terrains. (Note the standard deviations overlap appears as purple.) Both controllers exhibit variation between steps on a given terrain and variation between terrains, however the variation in the force sensing ID-CLF-QP leads to better tracking performance. Fig. \ref{fig:outputs} b. depicts the tracking performance with the mean. The human motion capture data used to generate the desired trajectory is also depicted. 
The swing tracking is similar for all controllers, so only one example is shown. The force sensing ID-CLF-QP achieves better tracking in stance than the ID-CLF-QP without force sensors on all terrains. Table \ref{tab:terrains} shows the root mean square error (RMSE) of the force sensing ID-CLF-QP (``Sensor'') over 11 step cycles is lower than the ID-CLF-QP with no force sensors (``No Sensor'') in stance for all four terrains.

Fig. \ref{fig:subjects} shows that the torque and tracking results for the four controllers for 2 subjects. The stance tracking results show better tracking performance from the force sensing ID-CLF-QP for both subjects compared to the 3 other controllers. Table \ref{tab:subjects} shows the RMSE of the tracking performance over 11 step cycles where the force sensing ID-CLF-QP has the lowest RMSE out of all of the controllers in stance for both subjects. The greater difference between RMSE values for the stance phase makes sense since this is when the prosthesis undergoes loading from the user and our controllers' vary in their response to these interaction forces. The force sensing ID-CLF-QP's better tracking for both subjects without tuning in between suggests this control method is more transferrable between subjects, meaning it could lend a method that works for multiple subjects without hours of expert tuning for each subject, as is currently required by impedance control methods \cite{AnnSimonConfig5}. Future work will test this control method on more subjects to further investigate its transferability between subjects.

These results show the force sensing ID-CLF-QP can achieve better tracking utilizing force sensors than without, suggesting that accounting for the forces in the dynamics allows this model-dependent controller to respond to its real-time loading conditions to achieve good tracking. 
This also suggests the improvement in model accuracy allows this model-dependent controller to better capture the nonlinearities of a trajectory, motivating the use of model-dependent prosthesis controllers to achieve more dynamic behaviors.
This motivates further study of model-dependent prosthesis control methods to assess if the model-dependence can yield a more transferable method between prosthetic devices and users, requiring less tuning and yielding formal guarantees of stability.

\section{Conclusion and Future Work} \label{sec:Conclusion}
This work achieved the first experimental realization of a model-dependent prosthesis knee controller in both stance and swing using real-time in-the-loop force measurements to complete the dynamics using an insole pressure sensor, load cell, and IMU resulting in stable human-prosthesis walking. By directly measuring the forces from the human and ground with the load cell and pressure sensor, we enable the prosthesis to account for its real-world conditions. These sensing methods increase the validity of our human-prosthesis stability guarantees and could empower a variety of amputees to walk in varying ways across changing terrain.

This controller achieved better tracking on 4 types of terrain compared to its counterpart without force sensors. Additionally this controller outperformed its counterpart with force estimation and a traditional PD controller on 2 subjects. These results demonstrate the controller's robustness and ability to adapt to varying external forces from the user and the ground. Its improved performance on 2 subjects without tuning in between suggests the real-time force response could replace the need to tune many parameters for every user and behavior, as is required in typical impedance prosthesis control methods \cite{AnnSimonConfig5}. Since this control method relies on real-time dynamic force sensing, as opposed to static tuned parameters, to respond to the forces induced by the user and the terrain, it could provide a more transferable method between users and behaviors, reducing the time amputees spend in a tuning session.

Future work will test more subjects, including an amputee subject, to further assess this controller's ability to achieve improved tracking performance while undergoing varying external forces. 
This controller will be applied to the ankle in addition to the knee for a more complex multi-contact gait \cite{zhao2016multi} that emulates human heel-toe roll, to achieve a more natural and efficient gait.
The CoP measurement from the pressure sensor enables use of CoP as a phase variable \cite{VirtConsCtrlProst}
and could be used as a guard to determine transitions in multi-domain multi-contact walking \cite{zhao2016multi}.

\bibliographystyle{IEEEtran}
\bibliography{bibliography}

\end{document}